%% file: dmlr.tex
\documentclass[twoside,11pt]{article}

\usepackage{blindtext}

\usepackage[abbrvbib, nohyperref,preprint]{dmlr2e}

\usepackage[utf8]{inputenc} 
\usepackage[T1]{fontenc}    
\usepackage{hyperref}       
\usepackage{url}            
\usepackage{booktabs}       
\usepackage{amsfonts}       
\usepackage{nicefrac}       
\usepackage{microtype}      
\usepackage{xcolor}         

\usepackage{amsmath}
\usepackage{mathtools}

\usepackage{subcaption}

\usepackage{algorithm}
\usepackage{algorithmic}

\usepackage{listings}

\usepackage{setspace}
\usepackage{threeparttable}
\usepackage{dsfont}

\usepackage[capitalize,noabbrev]{cleveref}

\usepackage{lastpage}
\dmlrheading{23}{2023}{1-\pageref{LastPage}}{9/23; Revised 10/23}{11/23}{21-0000}{Author One and Author Two} 
\def\openreview{\url{https://openreview.net/forum?id=XXXX}} 

\ShortHeadings{ShIOEnv: Synthesizing Bash Shell Inputs with Execution Artifacts}{Ragsdale and Boppana}
\firstpageno{1}

\begin{document}

\title{ShIOEnv: An Environment for Synthesizing Bash Shell Inputs with Execution Artifacts}

\author{\name Jarrod Ragsdale \email jarrod.ragsdale@my.utsa.edu \\
       \addr Department of Computer Science\\
       University of Texas at San Antonio \\
       San Antonio, TX, 78249-1644, USA
       \AND
       \name Rajendra Boppana \email rajendra.boppana@utsa.edu \\
       \addr Department of Computer Science\\
       University of Texas at San Antonio \\
       San Antonio, TX, 78249-1644, USA}

\editor{My editor}

\maketitle

\begin{abstract}
We introduce ShIOEnv, a Gymnasium-compatible Bash environment for synthesizing execution-annotated shell input traces. Datasets containing such traces are useful for training models to perform execution-free shell simulation, which in turn is useful for deploying honeypots for cyber deception. Existing datasets, such as NL2CMD and other Bash command corpora, do not include complete execution artifacts (stdout, stderr, exit status, and changes to the system's state) associated with a command's execution. ShIOEnv synthesizes shell inputs, executes them in a controlled system context, and records stdout, stderr, exit status, and structured filesystem and environment patches. Unconstrained (random) generation of shell inputs (command and arguments) results in multiple instances of arguments that are mostly redundant or result in execution errors. We introduce grammar-constrained synthesis to generate syntactically valid arguments with high input irreducibility, which we estimate through self-ablative contrasts of execution artifacts. We prepare 1.6M samples using grammar-constrained and unconstrained synthesis and collect their execution artifacts for 86 Linux/Bash utilities. Using this dataset, we train small language models (CodeT5) and evaluate their accuracy in predicting the actual execution behavior of attacker-sourced Bash commands. These models outperform commonly used rule-based simulation (Cowrie) and prompted large language models by up to 14.2 pp in output edit similarity (O-ES). We enhance the NL2CMD dataset by constructing 500K samples executable in a grounded system context along with their complete execution artifacts.
We empirically demonstrate that models trained on grammar-constrained samples outperform those trained on NL2CMD samples by 22.5 pp in O-ES. ShIOEnv, command grammars, and 2.1M samples used in our primary evaluations are released for use by other researchers.
\end{abstract}


\input{content/1_intro}
\input{content/2_background}
\input{content/3_method}
\input{content/4_eval}
\input{content/5_discussion}
\input{content/6_conclusion}


\impact{
This work introduces ShIOEnv, a data-generation environment and execution-annotated Bash dataset for studying execution-free command-line simulation.
The primary positive impact is the development of safer, more reproducible methods for modeling command behavior without executing commands at inference time.
Such models can support cyber-deception systems, command-line assistants, and evaluation frameworks in which direct command execution may be unsafe, unavailable, or undesirable.
By releasing the dataset, command grammars, and execution artifacts, this work supports reproducible research on system-grounded execution-free simulation.

The work introduces potential risks.
The dataset contains command strings, system outputs, and behavior annotations that may help models generate more plausible command-line interactions that could be misused to improve automated reconnaissance, produce more realistic interaction traces, or support deceptive command-line emulation.
The produced dataset is intended for execution-free simulation, benchmark construction, data-quality analysis, and cyber-deception research.

The dataset does not intentionally contain personal data.
Because samples are grounded in a constructed system context, records may include synthetic file names, paths, environment variables, and command outputs that reflect the curated execution environment.
Researchers using this resource should treat these artifacts as system-specific annotations and should not infer human behavior, operational prevalence, or demographic properties from the command distribution.
The authors will maintain the public dataset release, document known limitations, and update the dataset card if substantive issues are identified.
}

\section*{Data Availability and Reproducibility Statement}

ShIOEnv and the produced command grammars are available at \url{https://github.com/synlab-jragsdale/ShIOEnv}.
We release 2.1M generated execution-annotated samples available at \url{https://huggingface.co/datasets/jragsdale1/ShIO-bash-26.1}.
The released artifacts include 500K UCS samples, 1.1M GCS samples, and 500K NL2CMD samples synthesized from UCS and GCS traces, as well as evaluation splits and scripts required to reproduce the experiments reported in this paper.
A supplementary dataset documentation PDF provides the dataset card, datasheet, intended-use statement, responsible-use considerations, licensing terms, hosting plan, maintenance plan, and versioning information.


\acks{
The research was sponsored by the Army Research Laboratory, accomplished under Cooperative Agreement Number W911NF-24-2-0180, and conducted in the Systems and Networks (SyN) Lab at The University of Texas at San Antonio using the Cyber Deception Testbed, funded by the Army Research Office under grant number W911NF-21-1-0188 and the National Security Agency through the awards H98230-21-1-0171 and H98230-20-1-0392. The views and conclusions contained in this document are those of the authors and should not be interpreted as representing the official policies, either expressed or implied, of the Army Research Laboratory or the U.S. Government.
}

\vskip 0.2in
\bibliography{references}

\newpage
\appendix
\input{content/7_appendix}

\end{document}

%% file: content/1_intro.tex
\section{Introduction}
\label{sec:intro}
Execution-free shell simulation is the task of predicting the execution behavior of a shell input, consisting of a command and optional arguments, within a specified system context.
The target behavior includes execution artifacts, such as terminal output, exit status, and changes to system state that may affect subsequent inputs.
The system context comprises execution-relevant state that affects command behavior, including the filesystem, environment variables, permissions, and user identity.
Execution-free simulation may be useful in settings where direct shell input execution is unsafe, costly, or undesirable.
Examples include cyber-deception systems that emulate attacker-facing shell interactions and command-line agents that use terminal state or execution feedback when selecting subsequent actions \citep{sladic2024llm,ubuntuworld,intercode_2024}.

Preparing models for this task requires a diverse range of shell inputs paired with their execution behavior in a system context.
Existing public resources provide only partial support for this supervision regime.
\cite{lin2018nl2bash} and \cite{fu2023nl2cmd} provide datasets that primarily support natural-language-to-command translation; cybersecurity training logs provide observed command sequences and metadata; and recent security datasets synthesize command-like activity for detection tasks \cite{svabensky2021dataset, Trizna_SLP_2022}.
These datasets do not, by themselves, provide complete execution artifacts for commands executed under a shared, reproducible system context

To address these gaps, we introduce the shell input-output environment (ShIOEnv), a Gymnasium-compatible data-generation environment for collecting execution behaviors for given commands in grounded system contexts.
ShIOEnv casts shell input synthesis as a Markov decision process in which an agent incrementally constructs a shell input through argument-level actions.
When construction terminates, ShIOEnv executes the completed input and records the resulting execution behavior.

Shell inputs collapse into common execution behavior patterns in which invalid or redundant portions of the input can obscure the effect of individual arguments.
As a consequence, syntactic validity alone is insufficient as a data-quality criterion for execution-annotated samples.
We define behavioral irreducibility as a measurement of argument redundancy with respect to the final observed execution behavior.
We perform this measurement via contrasts of execution behaviors of argument-ablated inputs.
An input is considered irreducible if no such ablations produce the same execution artifacts.
However, constructing arguments in a command-agnostic manner, which we call unconstrained synthesis (UCS), yields low-irreducibility inputs in which execution behavior is dominated by invalid or redundant arguments.
We therefore introduce grammar-constrained synthesis (GCS), in which a command-specific grammar dictates a derivation for producing each argument.

This work treats shell-command corpus construction as a data-centric machine-learning problem, with the central question being how the data-generation process and irreducibility criteria affect the utility of execution-annotated shell inputs.
We generate 1.6M execution-annotated samples using grammar-constrained and unconstrained synthesis.
We also construct 500K execution-annotated samples from NL2CMD commands adapted to a specified system context.
Together, these data sources support controlled comparisons among synthesis methods (GCS and UCS) and adapted input-only command corpora from the literature (NL2CMD).
We evaluate the resulting datasets by training models for execution-free shell simulation and using model performance as a measure of dataset utility.
Models trained on GCS data outperform models trained on unconstrained inputs, adapted NL2CMD inputs, and external execution-free baselines such as Cowrie \citep{cowrie}.
Moreover, grammar-constrained samples demonstrate greater training efficiency, outperforming models trained on five times as many unconstrained samples.

To summarize, we make the following contributions:
\begin{itemize}
    \item We introduce ShIOEnv, a data-generation environment for constructing shell inputs and their execution artifacts for a given system context and for efficiently evaluating argument redundancy. Each sample links a shell input to its execution artifacts.
    \item We introduce grammar-constrained synthesis as a data-generation intervention for shell-command corpora. Command-specific grammars, derived from the man pages, restrict argument generation to context-admissible regions of the input space, reducing degenerate execution behavior and improving the mapping of shell arguments to the resulting execution behaviors.
    \item We curate and release 2.1M execution-annotated samples spanning 86 Linux commands, together with command grammars, execution artifacts, and dataset schema information required for reproducible use.
    \item We evaluate the generated data by assessing its ability to train models to simulate the specified system's execution behavior, empirically demonstrating improved performance over existing datasets and execution-free approaches.
\end{itemize}

The remainder of the paper is organized as follows.
\Cref{sec:prelim} introduces preliminaries and related work in shell modeling, synthesis, and executable environments.
\Cref{sec:methods} presents ShIOEnv, irreducibility estimation, and grammar-constrained synthesis.
\Cref{sec:eval} evaluates estimator fidelity, the distributional effect of grammar constraints, and downstream execution-free modeling utility.
\Cref{sec:discussion} discusses the implications of the results and remaining limitations.
\Cref{sec:conclusion} summarizes the findings and outlines future work.
\Cref{sec:append_dataset} provides supplemental details on the execution environment and generated datasets.

%% file: content/2_background.tex
\section{Background}
\label{sec:prelim}

\subsection{Preliminaries}
ShIOEnv models shell input synthesis as a sequential decision process in which an agent incrementally constructs a shell input for execution in a controlled system context.
A shell input consists of a command string (e.g., \texttt{find}, \texttt{ls}) followed by optional arguments (e.g., \texttt{-l}, \texttt{--list}).
Our formulation consists of three complementary parts: Markov decision processes to form the basis of argument-level construction, context-free grammars to define a command-admissible argument space, and temporal abstraction to treat variable-length grammar derivations as single argument-level actions.

A Markov decision process is a tuple $M=\langle \mathcal{S}, \mathcal{A}, p, r\rangle$, where $\mathcal{S}$ is a set of states, $\mathcal{A}$ is a set of actions, $p(s' \mid s,a)$ is a transition function, and $r(s,a)$ is a reward function \citep{puterman2014markov}.
In ShIOEnv, states represent partially constructed command-argument sequences, actions append completed arguments or terminate construction, and terminal transitions execute the completed input in the target system context.

A context-free grammar is a tuple $G=\langle N,\Sigma,\sigma,P\rangle$, where $N$ is a set of nonterminals, $\Sigma$ is a set of terminals, $\sigma\in N$ is a start symbol, and $P$ is a set of productions defining nonterminal expansions $n\rightarrow (N\cup \Sigma)^*$ for all $n \in N$ \citep{hopcroft2001cfg}.
A derivation expands nonterminals until only terminals remain.
The set of terminal sequences derivable from $\sigma$ defines the language $L(G)$.
ShIOEnv uses the notation $G_{c,x}=\langle N_c,\Sigma_{c,x},\sigma_c,P_c\rangle$ to denote the grammar used to construct inputs for the active command $c$ in a system context $x$.

Because a complete shell argument may require a variable-length grammar derivation, ShIOEnv treats each derivation as a temporally abstracted action.
This is consistent with semi-Markov decision processes, where a high-level action may span multiple lower-level steps before returning control to the decision process \citep{sutton1999options}.
Thus, ShIOEnv preserves argument-level decision epochs while using grammar expansion to construct admissible completed arguments.

\subsection{Related Work}
\paragraph{Shell Modeling}
Execution-free shell simulation is relevant in settings where direct command execution is unsafe, such as cyber-deception systems, where attackers could escape containment and abuse the resources used for direct execution \citep{sladic2024llm, otal2024llm}.
A more benign setting for execution-free simulation is found in command-line agents that must reason about terminal behavior before taking action \citep{agarwal2020nlc2cmd, intercode_2024}.
Prior work has used model-based methods to generate shell responses for interactive deception and terminal emulation \citep{sladic2024llm, otal2024llm,ragsdale2024evaluating}.
These systems primarily target plausible or engaging shell interaction rather than supervised prediction of exact executor-specific artifacts, such as stdout, stderr, exit status, and persistent state changes.

The command token, subsequent arguments, and the system context all jointly affect execution behavior.
A training example for this setting should therefore associate an input with the context in which it is executed in addition to the resulting text generation (stdout and stderr) and exit code.
Datasets containing only one part of this association are insufficient, as they do not provide the requisite mapping for training.
This motivates the development of shell input corpora annotated with their execution artifacts, in which observable responses and structured state transitions serve as supervised targets.

\paragraph{Shell Datasets and Command Synthesis}
Existing public shell datasets only partially support system-specific shell simulation.
The Linux Plan Corpus provides a limited input-to-behavior mapping, cybersecurity-training datasets record shell commands and execution metadata from participants, and NL2Bash and NL2CMD primarily support natural-language-to-command translation \citep{linuxplancorpus,svabensky2021dataset,lin2018nl2bash,fu2023nl2cmd}.
These resources are useful for studying command usage or command generation, but they are not designed to provide complete execution artifacts for commands executed under a shared, reproducible system context.

Command synthesis can be viewed as an instance of syntax-guided synthesis, where programs are generated from a grammar-constrained space to satisfy a given specification \citep{alur2013sygus}.
In this formulation, the grammar restricts the syntactic space of admissible candidates and functions \citep{alur2013sygus}.
Prior work has also used grammar-guided methods for program synthesis and scalable data generation \citep{Parsert_Polgreen_2024}.
In our setting, command-specific context-free grammars encode sequential production steps to produce an admissible argument.
We adopt this grammar-guided view, but use derivations to generate grammar-admissible shell arguments for execution and behavioral measurement rather than to synthesize programs against a logical specification.

The closest published work to the synthesis component of ShIOEnv is NL2CMD.
NL2CMD extracts utility syntax, flags, and argument categories from man pages to generate commands, covering 38 utilities from the original NL2Bash distribution \citep{fu2023nl2cmd}.
Bash commands are synthesized from these extracted elements to be executed and later paired with natural-language descriptions, yielding a dataset for natural-language-to-command translation.
Its execution step functions as a validation filter: placeholder arguments are instantiated, commands are executed in a virtual machine, and commands that fail to terminate with exit status zero are discarded.
The resulting corpus enables a scalable mechanism for generating shell inputs.
However, NL2CMD differs from ShIOEnv in objective and annotation scope.
ShIOEnv employs command grammars to constrain and direct the generation of arguments over a variable number of production steps, considering execution as a supervisory source rather than as a final preservation criterion.
Furthermore, each retained sample is accompanied by execution artifacts that are crucial for shell modeling but not required for natural-language-to-command translation.

\paragraph{Executable Environments}
Executable command-line environments support agents that interact with terminal-like systems and condition subsequent actions on observed feedback. 
InterCode formulates interactive coding as a reinforcement-learning environment with code actions and execution-feedback observations, including Bash-based tasks \citep{intercode_2024}.
UbuntuWorld uses the Bash terminal as a simulator for automated Ubuntu technical-support agents, while CLAI exposes the command line through a generic sense-act interface for AI agents \citep{ubuntuworld,clai}.
These environments primarily target task completion, troubleshooting, or interactive agent evaluation.
They do not report systematic exploration of command argument spaces, construction of execution-annotated shell corpora, or measurement of argument-level behavioral necessity.
Because these environments and corpora are not designed around randomized or grammar-constrained argument synthesis, they do not directly address the argument-redundancy problem studied in this work.
Consequently, existing resources leave open the problem of constructing shell-input corpora that jointly support controlled argument-space exploration, system-specific execution annotation, and empirical measurement of whether individual arguments are behaviorally necessary.

%% file: content/3_method.tex

\section{ShIOEnv}
\label{sec:methods}
We present ShIOEnv as a data-generation environment for constructing execution-annotated Bash shell inputs.
Given a target system context, ShIOEnv synthesizes shell inputs, executes completed inputs in that context, and records execution artifacts.
A system context specifies the filesystem, environment variables, user identity, permissions, and other execution-relevant state used by the executor.

ShIOEnv separates input construction from execution, allowing the environment to expose a controlled argument-level decision process while deferring system effects, execution artifacts, and contrastive ablations (to assess irreducibility) until a complete shell input has been formed.
During construction, the synthesizing agent incrementally builds a shell input by appending completed arguments to a command token.
When construction terminates, the completed input is executed and its irreducibility measured in the specified system context.

\Cref{subsec:env_formulation} presents an argument-level MDP abstraction for describing the construction process.
This abstraction intentionally omits low-level interface details to improve clarity.
The Gymnasium-compatible operational implementation used in our evaluation is provided in \Cref{sec:append_env}.
\Cref{subsec:rew_signal} defines behavioral irreducibility as the terminal signal used to score completed inputs.
\Cref{subsec:gcs} defines grammar-constrained synthesis, which restricts argument construction to command-specific action regions by treating variable-length grammar derivations as temporally abstracted argument actions.

\subsection{Environment Formulation}
\label{subsec:env_formulation}

A shell input consists of a command token followed by an optional sequence of arguments whose execution behavior is evaluated in the specified system context.
For example, the default behavior of the disk listing command \texttt{df} is altered by arguments \texttt{-a} to show all candidates, \texttt{-{}-output=source} to only show the source column, and \texttt{-B 512K} to print sizes as 512 kilobyte blocks.
Let $x \in \mathcal{X}$ denote the instantiated system context.
Let $c \in C = \{\texttt{df}, \texttt{ls}, \dots\}$ denote a command token and $a_1, \dots, a_n$ denote each following argument action.
Let $\mathcal{V} \subseteq \Sigma^*$ denote the set of complete argument strings over vocabulary $\Sigma$.

\begin{figure*}[htbp]
  \centering
  \includegraphics[width=0.95\linewidth]{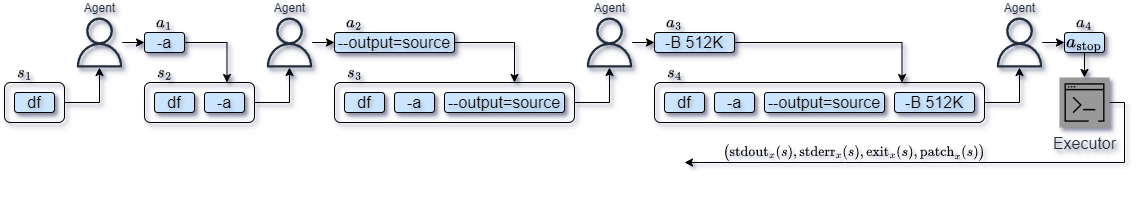}
  \caption{Synthesis of shell input containing a \texttt{df} command and argument actions. Execution artifacts are collected upon completion. }
  \label{fig:shio_state_action}
\end{figure*}

Using the MDP tuple $\langle \mathcal{S},\mathcal{A},p,r \rangle$ \citep{puterman2014markov}, we define construction states as ordered command-argument sequences:
\[
\mathcal{S}
=
\{(c,a_1,\ldots,a_n) \mid n \geq 0,\ c \in C,\ a_i \in \mathcal{V}\}.
\]
The construction state represents a shell input built up to that point.
The system context $x$ is not included in this state because it is specific to the environment used during a rollout.
The action set contains complete argument strings and a special stopping action:
\[
\mathcal{A}
=
\mathcal{V} \cup \{a_{\mathrm{stop}}\}.
\]
The transition function is deterministic for every construction state.
If $a \neq a_{\mathrm{stop}}$, the action appends $a$ to the current command-argument sequence.
If $a = a_{\mathrm{stop}}$, the environment transitions to a terminal state and executes the completed input in the target system context, producing the execution behavior
\[
E_x(s)
=
\bigl(
\operatorname{stdout}_x(s),
\operatorname{stderr}_x(s),
\operatorname{exit}_x(s),
\operatorname{patch}_x(s)
\bigr),
\]
denoting the execution artifact produced by executing $s$ in context $x$.
The first three components capture observable terminal feedback exposed to the user.
These components characterize the command's immediate response, though they do not capture persistent effects on the execution context.
The final component, $\operatorname{patch}_x(s)$, records state changes as a structured patch within the system context, capturing execution effects that would otherwise be hidden at the current step.
Context patches are retained because later inputs in an interactive shell session may depend on effects such as file creation, removal, permission changes, or changes to the working directory.
Observable records provide response-level supervision.
Context patches provide state-transition supervision for system-specific shell simulation.

The synthesis process is illustrated in \Cref{fig:shio_state_action}, where argument actions synthesized by the agent while interacting with the environment are appended to the construction state until $a_{\mathrm{stop}}$ signals that the input is ready to be executed and artifacts are collected.
We calculate the reward using the irreducibility measurement discussed in \Cref{subsec:rew_signal}.

\subsection{Behavioral Irreducibility}
\label{subsec:rew_signal}
Given the execution artifact $E_x(s)$ defined above, behavioral irreducibility estimates the extent to which the artifact depends on the arguments in $s$.
Intuitively, an input is irreducible if the removal of any number of arguments results in a change in $E_x$.
Conversely, an input is reducible if one or more arguments can be removed without altering the observed execution artifacts.
For example, removing \texttt{-s} or \texttt{-r} from the shell input \texttt{uname -a -s -r} would have no effect on execution, as their behavior is dominated by \texttt{-a}.
We denote an input as fully irreducible when every proper retained-argument contrast changes the artifact.

Let $N=\{1,2,\ldots,n\}$ denote the ordered set of argument indices for a completed input $s=(c,a_1,\ldots,a_n)$.
For any retained-index set $I \subseteq N$, let
\[
s[I]=(c,a_{I_1},\ldots,a_{I_{|I|}})
\]
denote an argument-ablated input formed by removing any arguments at indices not in $I$.
We contrast the $E_x$ of the full input against the $E_x$ of each ablated input.
Let $\Delta_x(s,s[I]) \in \{0,1\}$ be an indicator function returning $1$ when $E_x(s) \neq E_x(s[I])$ and $0$ otherwise.
Two artifacts are treated as equivalent when their exit codes and context patches match and when the normalized edit similarity of stdout and stderr exceeds the noise threshold described in \Cref{sec:append_env}.

Not all contrasts are equally informative.
A less extensively ablated input is likely to correspond more closely with the candidate input than one with a greater number of arguments removed.
We therefore weight each execution contrast by the number of retained arguments measured by $\lvert s[I] \rvert$.
Let $\mathcal{O}_n = \{I \subseteq N \mid |I| < n\}$ be the set of proper retained-index sets.
For inputs with at least one argument, the irreducibility score is given as
\[
R_x^\star(s)
=
\frac{
\sum_{I \in \mathcal{O}_n} \lvert s[I] \rvert\,\Delta_x\bigl(s,s[I]\bigr)
}{
\sum_{I \in \mathcal{O}_n} \lvert s[I] \rvert
}, \qquad \lvert \mathcal{O}_n \rvert \geq 1.
\]
For inputs with no arguments, we set $R_x^\star(s)=1$, since no argument ablations are possible.
The weight $\lvert s[I] \rvert$ upweights near-neighbor ablations and downweights heavily pruned comparisons.
The normalization term makes the score comparable across input lengths.
This definition captures higher-order argument interactions that would be missed by considering only single-argument removals.

Because exhaustive irreducibility requires evaluating all properly retained subsets of arguments, its computational cost increases exponentially with the number of arguments.
We therefore approximate $R_x^\star(s)$ using a fixed-budget Monte Carlo estimator.
Let
\[
B=\min(M,\lvert\mathcal{O}_n\rvert)
\]
be the number of ablations evaluated, and let $\{I_1,\ldots,I_B\} \subseteq \mathcal{O}_n$ be retained-index sets drawn uniformly without replacement, using all retained-index sets when $B=\lvert\mathcal{O}_n\rvert$.
The estimator is given as
\[
\widehat{R}_{M,x}(s)
=
\frac{
\sum_{m=1}^{B} \lvert s[I] \rvert,\Delta_x\bigl(s,s[I_m]\bigr)
}{
\sum_{m=1}^{B} \lvert s[I] \rvert
}.
\]
In ShIOEnv, $\widehat{R}_{M,x}(s)$ is returned as the terminal reward for completed inputs.
The evaluation uses this terminal reward as a data-quality measurement and filtering signal rather than optimizing a learned sampling policy against it.
When the system context is fixed and known, we omit the subscript $x$, written as $R^\star$ or $\widehat{R}_M$.

\subsection{Grammar-Constrained Synthesis}
\label{subsec:gcs}

The unconstrained action space contains argument strings that are not valid for every command.
For example, \texttt{-{}-output=source} is a valid argument \texttt{df}, but not for \texttt{find}, \texttt{grep}, or \texttt{uname}.
Consequently, unconstrained (random) synthesis can produce shell inputs that collapse into repeated usage errors.
We address this problem by constraining synthesis via command-specific CFGs.

For command $c$ under system context $x$, let
\[
G_{c,x}
=
\langle N_c,\Sigma_{c,x},\sigma_c,P_c\rangle
\]
denote a context-free grammar for a command $c$ with terminals in the system context $x$.
The terminal alphabet $\Sigma_{c,x}$ includes context-dependent values drawn from $x$, such as filenames and environment variables.
The language of the grammar satisfies $L(G_{c,x}) \subseteq \mathcal{S}$, wherein a shell input is grammar-admissible for command $c$ in context $x$ if it belongs to $L(G_{c,x})$.

Let $\sigma_a \subseteq (N_c \cup \Sigma_{c,x})$ be the set of terminal and nonterminal tokens starting each independent argument.
We make this distinction in order to cleanly separate argument actions for ablations.

\begin{figure}[htb]
\centering    
\begin{subfigure}{0.48\textwidth}
\centering
  \includegraphics[width=\textwidth]{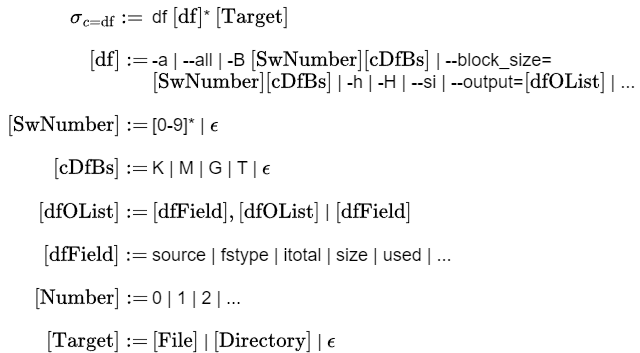}
    \label{fig:grammar_subfig}
\end{subfigure}
\hfill
\begin{subfigure}{0.48\textwidth}
\centering
  \includegraphics[width=\linewidth]{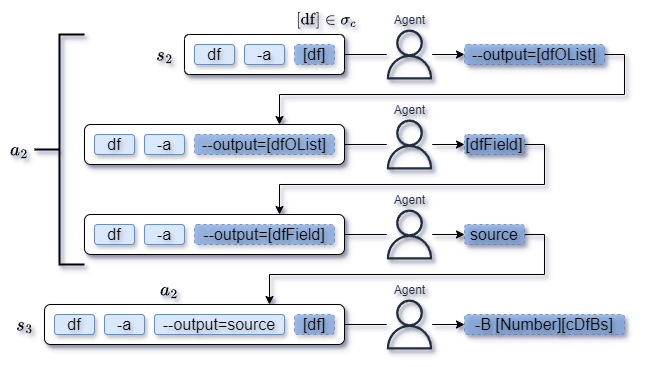}
  \label{fig:options_subfig}
\end{subfigure}
\caption{Expansion of \texttt{-{}-output=source} argument by \texttt{df} grammar productions.}
\label{fig:option_fig}
\end{figure}

For a construction state $s=(c,a_1,\ldots,a_n)$, an internal derivation constructs an argument action consisting of terminals found from expansions rooted at any $\sigma_a$.
We provide \Cref{fig:option_fig} to illustrate this process where $\sigma_a = \{\textrm{df}, \textrm{[df]}, \textrm{[Target]}\}$.
Productions are followed to expand a long-option nonterminal from $\textrm{[df]}$ into \texttt{-{}-output=source}.
The resulting terminal string \texttt{-{}-output=source} is emitted as one completed argument ($a_2$) action.
This process is performed in a single time step from the environment's perspective to preserve temporal consistency.

Internal derivations are not exposed as ShIOEnv states and do not create intermediate rewards or executions.
The grammar expansion can be interpreted as a semi-MDP over production steps \citep{sutton1999options}, while ShIOEnv retains decision epochs only at the argument level.
Grammar-constrained synthesis therefore operates without changing the argument-level construction semantics of the environment.

We compare grammar-constrained synthesis with unconstrained synthesis.
In unconstrained synthesis, inputs are produced in a command-agnostic manner (all command arguments considered).
In grammar-constrained synthesis, each emitted argument is generated by a derivation in $G_{c,x}$, so the action belongs to $L(G_{c,x})$ and is admissible for command $c$ in context $x$.
This comparison isolates the effect of command-specific grammar constraints on argument irreducibility and execution-free shell simulation.
As shown in \Cref{sec:eval}, grammar-constrained synthesis shifts the synthesized distribution toward higher irreducibility and improves execution-free modeling performance.
Irreducibility filtering provides an additional data-quality signal for evaluating shell inputs.

\begin{table}[htb]
    \centering
    \begin{threeparttable}
        \caption{Evaluated commands}
        \label{tab:cmds}
        \begin{tabular}{llll llll}
        \toprule
        arch & basename & cat & cd\tnote{1} & chgrp & chmod & chown & comm\tnote{2} \\
        cp & cpio\tnote{1,2} & curl\tnote{1} & cut\tnote{2} & date & dd & df & dir \\
        dirname & dmesg\tnote{1} & du & echo & env & ethtool\tnote{1} & export\tnote{1} & false \\
        file\tnote{1} & find\tnote{1} & fold\tnote{2} & free\tnote{1} & grep\tnote{1,2} & groups & gzip\tnote{2} & head\tnote{2} \\
        hostid & id & ip-addr\tnote{1} & ip-link\tnote{1} & ip-route\tnote{1} & last\tnote{1} & link & ln \\
        logname & ls & lsblk\tnote{1} & lscpu\tnote{1} & mkdir & mkfifo & mktemp & mv \\
        netstat\tnote{1} & nl\tnote{2} & nproc & od\tnote{2} & paste\tnote{2} & ping\tnote{1} & printenv & ps\tnote{1} \\
        pwd & readlink & rev\tnote{1,2} & rm & rmdir & shred & sort\tnote{2} & split\tnote{2} \\
        stat & tail\tnote{2} & tar\tnote{1} & tee\tnote{2} & touch & tr\tnote{2} & true & truncate \\
        ulimit\tnote{1} & uname & unexpand & uniq\tnote{2} & unlink & uptime & users & vdir \\
        w\tnote{1} & wc\tnote{2} & wget\tnote{1} & who & whoami & xargs\tnote{1,2} & & \\
        \bottomrule
        \end{tabular}
        \begin{tablenotes}
            \item [1] non-core utility.
            \item [2] Defined pipe behavior.
        \end{tablenotes}
    \end{threeparttable}
\end{table}

We construct command-specific grammars from 6,222 productions spanning 86 commands for grammar-constrained synthesis.
The supported commands are intended to provide a controlled evaluation space of the larger shell input space, covering common command-line behaviors, including filesystem inspection, file mutation, permission modification, text processing, environment inspection, system inspection, networking, compression, and command composition.
This scope supports controlled comparisons among unconstrained synthesis, grammar-constrained synthesis, and irreducibility filtering while preserving behavior types needed to evaluate stdout, stderr, exit status, and context-patch prediction.
The list of supported commands is provided in \Cref{tab:cmds}.
The grammar suite includes most GNU Coreutils commands and commonly installed utilities, such as \texttt{curl}, \texttt{ip}, and \texttt{wget}, to cover a broad range of expected behaviors in a Linux shell environment.
Each grammar includes constructions for inputs as part of a logical operation (\texttt{||, \&\&}), output redirection (\texttt{>, >\;>}), or an I/O pipe (\texttt{|}) for those with defined pipe usage.

%% file: content/4_eval.tex
\section{Evaluation}
\label{sec:eval}
We evaluate ShIOEnv-synthesized datasets by measuring the effect of grammar constraints on execution-free simulation.
Using datasets constructed under different synthesis regimes, we evaluate whether these mechanisms improve the performance of fine-tuned models predicting execution artifacts when conditioned on the shell input that produced them.
Finally, we slice each dataset by irreducibility to jointly measure the effects of grammar-constrained synthesis and sample irreducibility.

\subsection{Experimental Setup}
\label{sec:setup}
The system context for collecting execution artifacts is instantiated in a Firecracker microVM running Ubuntu 24.04 with 1 CPU core and 256 MB of memory \citep{firecracker}.
Four Dell R730s are used to create 200 Firecracker microVMs associated with a ShIOEnv environment
The microVM is snapshotted upon instantiation and reverted to its initial state after each execution.
This ensures a consistent operating environment for irreducibility calculations.
The host communicates with the microVM via a virtual socket created during boot using the Firecracker API.
This virtual socket enables communication between the host and the microVM for execution and for returning artifacts observed during execution.

We consider three data sources.
NL2CMD denotes commands adapted from the NL2CMD corpus by replacing generic placeholders with system-specific artifacts present in the target environment \citep{fu2023nl2cmd}.
UCS denotes unconstrained synthesis, in which arguments are sampled from all possible arguments in a command-agnostic manner.
GCS denotes grammar-constrained synthesis, in which each argument is generated from the active command grammar.

We evaluate the effectiveness of these data sources for preparing models to conditionally generate execution artifacts.
Specifically, we train encoder-decoder transformers initialized from CodeT5-base (220M parameters) \citep{wang2021codet5} on datasets that differ in data source, volume, system context, and irreducibility.

We include existing execution-free baselines in our evaluation to ground the performance of models trained on ShIOEnv datasets. 
Cowrie represents a rule-based SSH honeypot baseline \citep{cowrie}.
Its exposed filesystem and environment are modified to match the target execution context, thereby estimating the fidelity attainable with hand-specified interaction rules under the same environment.
We also evaluate GPT-4o-mini and GPT-4.1-mini as prompted, general-purpose language model baselines.
These models receive a serialized description of the system context and the input shell command, but they receive no task-specific post-training.
These models serve to separate the effect of task-specific, execution-annotated training data from that of general-purpose language-model prior knowledge.

We evaluate the performance of these methods using three complementary metrics.
Exact match (EM) measures the proportion of samples for which the predicted execution behavior is identical to the target behavior after canonicalization.
We define canonicalization as a deterministic normalization of execution artifacts that abstracts away incidental numeric values and formatting differences.
Specifically, each distinct numeric value is replaced by an ordered marker according to its first occurrence, so that the first distinct number is mapped to \texttt{<NUM\_0>}, the next distinct number to \texttt{<NUM\_1>}, and later occurrences of the same value receive the same marker.
Contiguous whitespace is similarly normalized to a standard whitespace representation.
This normalization is used to prevent incidental differences such as environment-dependent numeric values, identifiers, counters, and formatting variation from being scored as execution-behavior errors, while still preserving the structural pattern of repeated values.
Observable-output edit similarity (O-ES) measures canonicalized edit similarity between predicted and target observable outputs, including stdout and stderr.
Context-patch Jaccard similarity ($\Delta$-JS) measures overlap between predicted and target canonicalized patch operations.
Together, these metrics distinguish strict end-to-end correctness, partial observable-output fidelity, and latent state-transition fidelity.

All methods are evaluated on the same benchmark of 357 shell inputs drawn from honeypot logs \citep{ragsdale2024evaluating}.
Each benchmark input is executed in the same system context to obtain the target observable output and context patch.
Training samples are generated independently of this benchmark, avoiding data leakage between training and evaluation.

\subsection{Execution-Free Simulation Performance}
\label{sec:execsim}
We next evaluate the effectiveness of datasets constructed in ShIOEnv by comparing the performance of models trained on them for execution-free simulation.
In this task, a model receives a shell input and predicts the execution artifacts produced by the controlled executor for the user inputs described in \Cref{sec:setup}.
We compare the performance of CodeT5 models trained on samples from each data source in addition to existing baseline methods of execution-free simulation.

\begin{table}[htb]
    \centering
    \caption{Bash command execution-free simulation performance (UCS=Unconstrained sampling, GCS=Grammar-constrained sampling).}
    \label{tab:base_eval}
    \begin{tabular}{llll}
    \toprule
    Method & EM & O-ES & $\Delta$-JS \\
    \midrule
        Cowrie        & 0.165 & 0.345 & 0.248 \\
        GPT-4o-mini   & 0.171 & 0.582 & 0.164 \\
        GPT-4.1-mini  & 0.247 & 0.593 & 0.570 \\
      \midrule
        CodeT5 + NL2CMD & 0.241 & 0.510 & 0.488 \\
        CodeT5 + UCS  & 0.333 & 0.605 & 0.383 \\
        CodeT5 + GCS  & \textbf{0.457} & \textbf{0.735} & \textbf{0.578} \\
    \bottomrule
    \end{tabular}
\end{table}

The results in \Cref{tab:base_eval} indicate that CodeT5 trained on GCS samples achieves the strongest overall performance among the evaluated execution-free methods.
Relative to CodeT5 trained on UCS samples, GCS improves EM by 12.4 percentage points, O-ES by 13.0 percentage points, and $\Delta$-JS by 19.5 percentage points.
Relative to CodeT5 trained on adapted NL2CMD samples, GCS improves EM by 21.6 percentage points, O-ES by 22.5 percentage points, and $\Delta$-JS by 9.0 percentage points.
GCS also exceeds the strongest prompted language-model baseline in EM and O-ES, while achieving comparable context-patch fidelity to GPT-4.1-mini.
These results indicate that grammar-constrained data improves both observable-output modeling and state-transition modeling.

\begin{table}[htb]
    \centering
    \caption{CodeT5 performance by training dataset and volume.}
    \label{tab:data_vol}
    \begin{tabular}{l llll}
    \toprule
    Size & Dataset & EM & O-ES & $\Delta$-JS \\
    \midrule
    100K & NL2CMD & 0.171 & 0.401 & 0.402 \\
     & UCS  & 0.207 & 0.446 & 0.191 \\
     & GCS  & 0.353 & 0.661 & \textbf{0.582} \\
     \midrule

    300K & NL2CMD & 0.227 & 0.465 & 0.354 \\
     & UCS  & 0.286 & 0.529 & 0.268 \\
     & GCS  & 0.426 & 0.715 & 0.523 \\
     \midrule
    
    500K & NL2CMD & 0.241 & 0.510 & 0.488 \\
     & UCS  & 0.333 & 0.605 & 0.383 \\
     & GCS  & \textbf{0.457} & \textbf{0.735} & 0.578 \\
    \bottomrule
    \end{tabular}
\end{table}

We next vary the training-set size to determine whether GCS gives greater effective sample coverage. 
\Cref{tab:data_vol} shows that GCS remains superior under substantially smaller training budgets.
A model trained on 100K GCS samples exceeds models trained on 500K NL2CMD or 500K UCS samples across all three metrics.
This result suggests that grammar-constrained synthesis improves data efficiency by increasing the behavioral relevance of individual samples.
Increasing GCS volume from 100K to 500K further improves EM from 0.353 to 0.457 and O-ES from 0.661 to 0.735.
In contrast, $\Delta$-JS varies with data volume for NL2CMD and GCS, indicating that context-patch prediction is sensitive to the distribution of filesystem and environment effects represented in the training corpus.

\begin{table}[htb]
    \centering
    \caption{CodeT5 performance distribution across execution environments (OS + filesystem).}
    \label{tab:multi_sys}
    \begin{tabular}{l l l l}
    \toprule
    Dataset & EM & O-ES & $\Delta$-JS \\
    \midrule
        NL2CMD & 0.245$\pm$0.00 & 0.465$\pm$0.04 & 0.447$\pm$0.04 \\
        UCS  & 0.372$\pm$0.04 & 0.608$\pm$0.03 & 0.347$\pm$0.04  \\
        GCS  & 0.462$\pm$0.01 & 0.719$\pm$0.01 & 0.591$\pm$0.06  \\
    \bottomrule
    \end{tabular}
\end{table}

Finally, we evaluate whether the observed gains persist across different execution contexts.
\Cref{tab:multi_sys} reports mean and standard deviation over three environments that differ in operating-system image (Ubuntu 24.04, Ubuntu 22.04, Debian 12) and filesystem contents (\Cref{fig:fs_tree}).
GCS achieves the highest mean performance across all metrics, with 0.462 EM, 0.719 O-ES, and 0.591 $\Delta$-JS.
The small standard deviations for EM and O-ES indicate that grammar-constrained synthesis provides consistent supervision across executor instances.
The larger variation in $\Delta$-JS suggests that latent state-transition prediction is more dependent on backend-specific filesystem and environment artifacts than observable execution artifacts.

\subsection{Impact of Irreducibility on Performance}
We report the irreducibility profiles of UCS and GCS and determine their effects on fine-tuning models for execution-free simulation.
We first determine the minimum number of argument-ablated execution contrasts required to accurately approximate irreducibility.
Using this approximation, we assess the distributional changes to irreducibility induced by grammar-constrained synthesis.
Finally, we filter GCS and UCS datasets by irreducibility before fine-tuning.

\subsubsection{Irreducibility Approximation}
Exhaustive irreducibility requires executing each possible argument ablation, making the computation exponential in the number of arguments.
We therefore evaluate whether Monte Carlo estimation provides a sufficiently accurate approximation of irreducibility for dataset construction.
Accuracy is measured by taking the mean absolute error (MAE) of the true irreducibility ($R^*$) and the estimated irreducibility ($R_M$) for exponentially increasing values of $M$.

\begin{figure}[htb]
\centering
\begin{subfigure}{0.49\textwidth}
\centering
  \includegraphics[width=\textwidth]{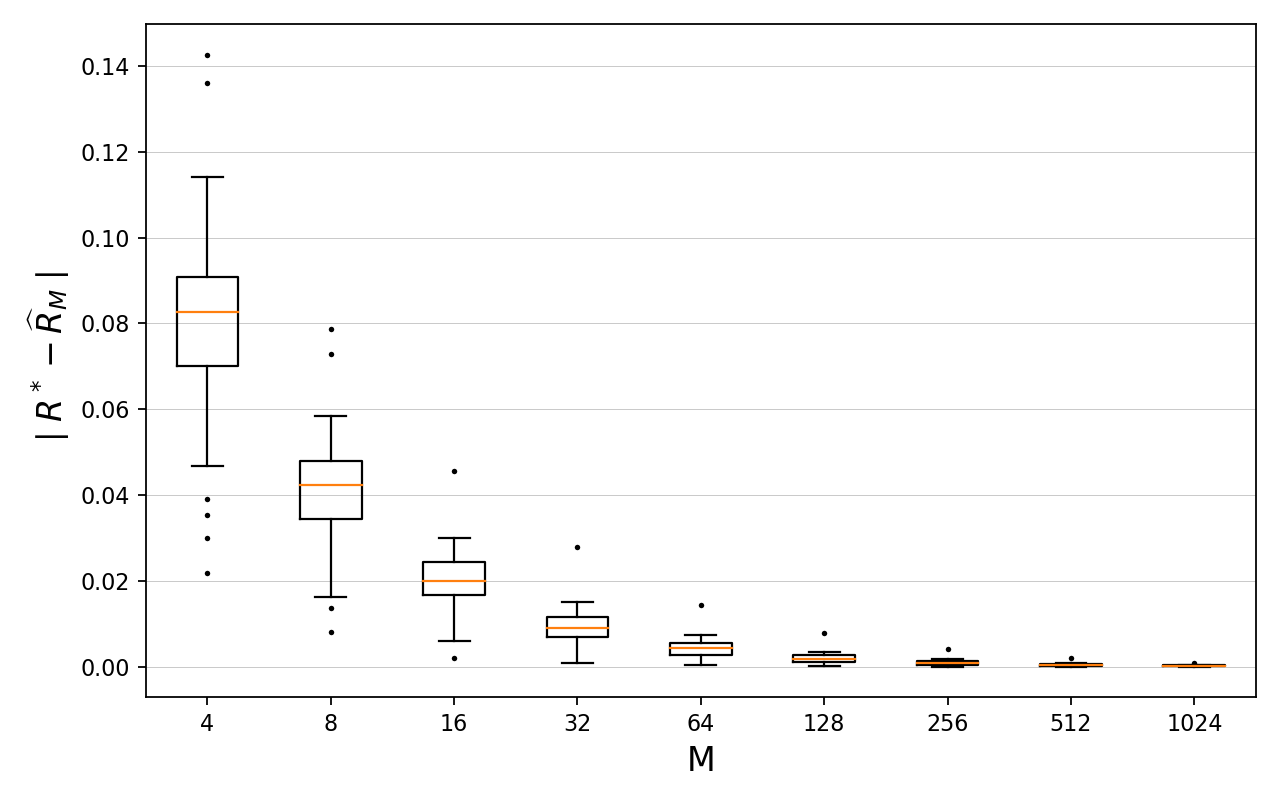}
    \caption{Command-level error}
  \label{fig:mae_cmd}
  
\end{subfigure}
\hfill
\begin{subfigure}{0.49\textwidth}
\centering
\includegraphics[width=\textwidth]{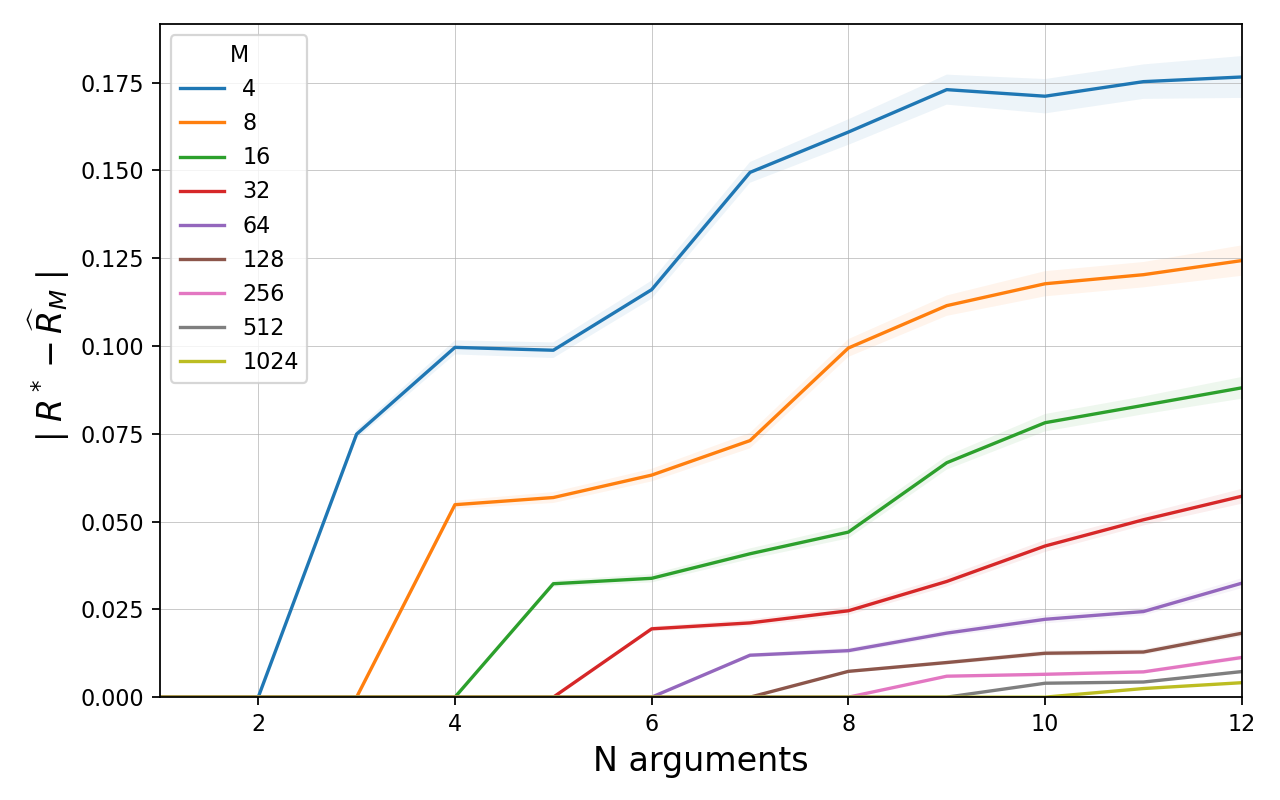}
      \caption{Error by argument length}
    \label{fig:mae_args}

\end{subfigure}
\caption{Approximation error of the budgeted irreducibility estimator. The left plot reports the command-level error distribution for each contrast budget. The right plot reports mean absolute error between $R^\star$ and $\widehat{R}_M$ by argument length. Shaded regions indicate 95\% confidence intervals.}
\label{fig:mae_budget}
\end{figure}

\Cref{fig:mae_cmd} shows the median command-aggregated MAE for increasing budgets.
The MAE decays rapidly as $M$ increases, with diminishing returns beyond moderate budgets.
\Cref{fig:mae_args} shows that approximation error increases with argument length for any fixed contrast budget.
This behavior is anticipated, as the quantity of argument ablations for a shell input corresponds to $2^n$, where $n$ denotes the number of arguments.
With a predetermined budget, the sampled fraction of ablations diminishes swiftly.
For 12-argument inputs, $M=64$ yields an MAE near 0.03 while sampling approximately 1.56\% of possible ablations.

Monte Carlo estimation is not an exact substitute for exhaustive ablation, especially for long inputs under very small budgets.
However, moderate budgets provide a practical approximation that preserves the irreducibility signal needed for filtering and analysis while avoiding an infeasible number of contrastive executions.
For these reasons, we use an ablation budget of $M=64$ for the large-scale synthesis needed to perform the following evaluations.

\subsubsection{Irreducibility Analysis of Synthesized Datasets}
\label{subsec:input_irreducibility}

\begin{figure}[htb]
\centering    
\begin{subfigure}{0.49\textwidth}
\centering
  \includegraphics[width=\textwidth]{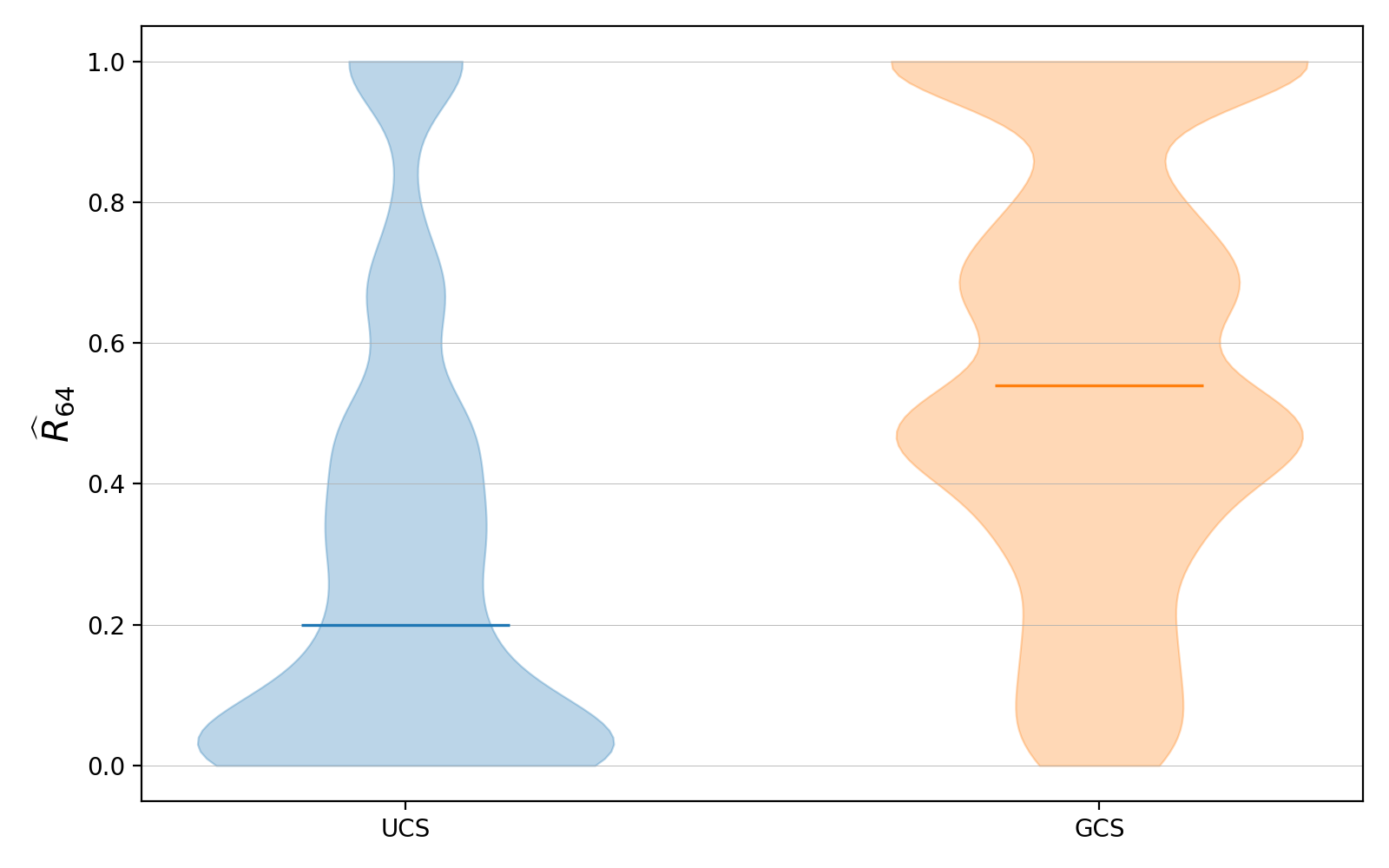}
  \label{fig:irred_violin}
\end{subfigure}
\hfill
\begin{subfigure}{0.49\textwidth}
\centering
  
  \includegraphics[width=\textwidth]{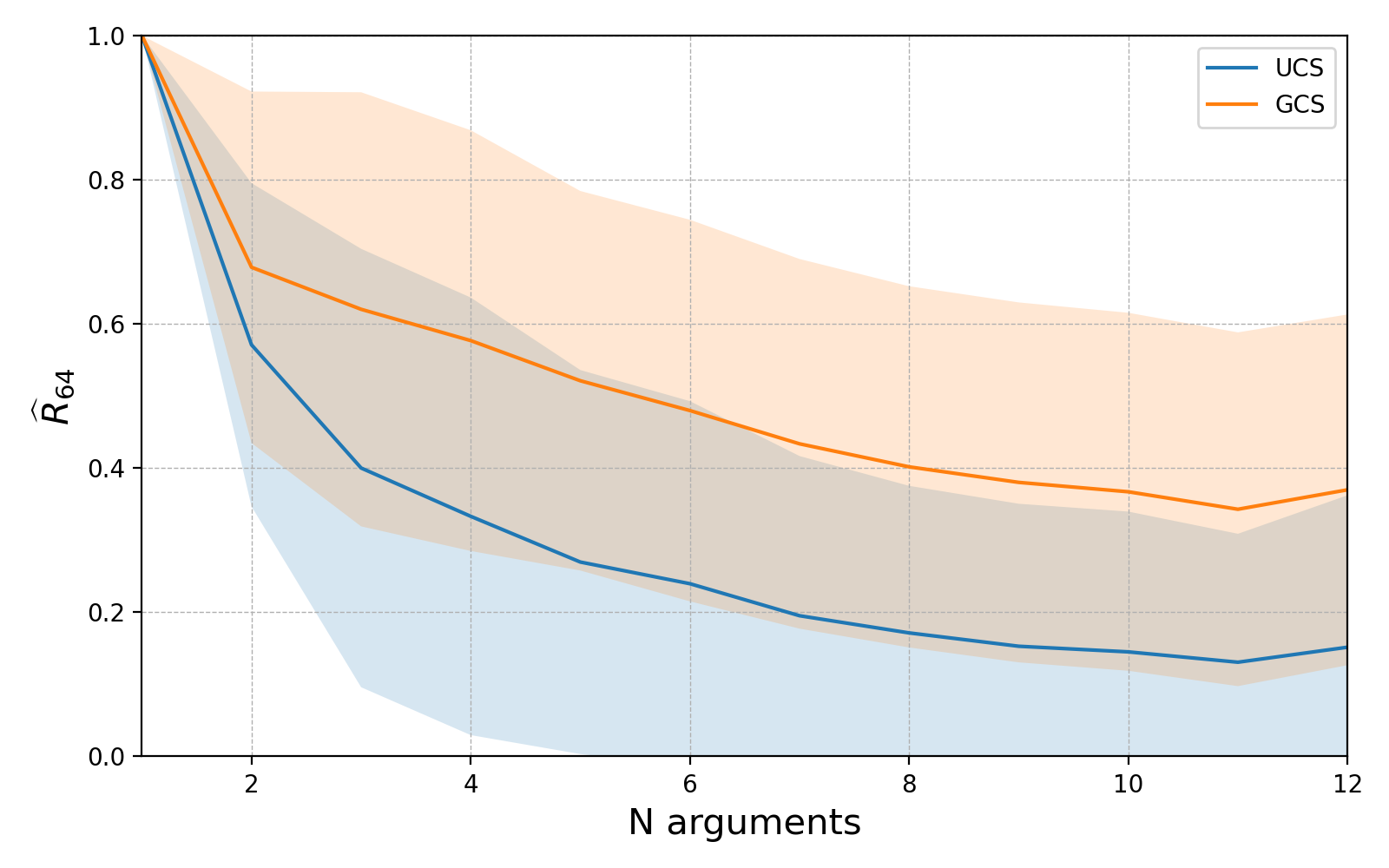}
    \label{fig:irred_by_length}
\end{subfigure}
\caption{Global distribution of irreducibility by synthesis method (left) and number of arguments (right) using $\widehat{R}_{M=64}$. UCS denotes unconstrained synthesis. GCS denotes grammar-constrained synthesis. Shaded regions indicate standard deviation.}
\label{fig:gcs_ucs_irred}
\end{figure}

We use the ablations described in \Cref{sec:execsim} to estimate the irreducibility of all synthesized shell inputs.
We plot this data in \Cref{fig:gcs_ucs_irred}, showing the global distribution of commands by irreducibility score for each dataset on the left and the irreducibility of inputs categorized by argument count on the right.
For each setting, we measure the irreducibility of the first 250K GCS and UCS shell inputs.
The global distribution in \Cref{fig:irred_violin} shows the proportional mass of shell inputs for UCS and GCS, plotted against irreducibility on the y-axis, with the mean indicated by the horizontal bar.

These distributions show that grammar constraints shift synthesized inputs toward higher irreducibility.
The proportion of samples with irreducibility greater than 0.5 increases from $24.90\%$ under UCS to $59.24\%$ under GCS.
UCS places substantial probability mass near low irreducibility values, while GCS shifts mass toward intermediate and high irreducibility values.
This result indicates that command-specific grammar constraints reduce invalid or degenerate argument combinations and increase the probability that emitted arguments contribute to execution behavior.

The argument-length distribution in \Cref{fig:irred_by_length} shows that irreducibility decreases as argument length increases under both synthesis regimes.
Longer inputs provide more opportunities for redundant arguments, option masking, repeated effects, and error-dominated behavior.
However, GCS maintains higher irreducibility across the full length range.
These results suggest that grammar constraints improve the behavioral support of the synthesized corpus by producing shell inputs whose arguments are more strongly coupled to execution behavior.

\subsubsection{Impact of Irreducibility on Execution-Free Simulation}

\begin{table}[htb]
    \centering
    \caption{CodeT5 execution-free modeling performance by dataset irreducibility. The threshold $\widehat{R}_{64}\geq 0.0$ denotes the unfiltered training set.}
    \label{tab:irred_filtering}
    \begin{tabular}{l l lll}
    \toprule
    Dataset & Filter & EM & O-ES & $\Delta$-JS \\
    \midrule
    UCS & $\widehat{R}_{64} < 0.5$ & 0.269 & 0.476 & 0.141 \\
    UCS & $\widehat{R}_{64} \geq 0.0$ & 0.333 & 0.605 & 0.383 \\
    UCS & $\widehat{R}_{64} \geq 0.5$ & 0.375 & 0.632 & 0.455 \\
    \midrule
    
    GCS & $\widehat{R}_{64} < 0.5$ & 0.359 & 0.616 & 0.392 \\
    GCS & $\widehat{R}_{64} \geq 0.0$ & 0.457 & \textbf{0.735} & 0.578 \\
    GCS & $\widehat{R}_{64} \geq 0.5$ & \textbf{0.482} & \textbf{0.735} & \textbf{0.631} \\
    \bottomrule
    \end{tabular}
\end{table}

The preceding results show that grammar constraints improve both irreducibility and downstream modeling performance.
We now test whether irreducibility itself functions as a data-quality signal within each synthesis regime.
This experiment compares models trained on three slices of each synthesized dataset.
The unfiltered slice has no criterion and is denoted by $\widehat{R}_{64} \geq 0.0$.
The high-irreducibility slice contains samples satisfying $\widehat{R}_{64} \geq 0.5$.
The low-irreducibility slice contains samples satisfying $\widehat{R}_{64} < 0.5$.
This design isolates the effect of retaining samples whose arguments are empirically coupled to execution behavior.

\Cref{tab:irred_filtering} shows that high-irreducibility filtering improves performance for both UCS and GCS.
For UCS, filtering to $\widehat{R}_{64} \geq 0.5$ improves EM from 0.333 to 0.375, O-ES from 0.605 to 0.632, and $\Delta$-JS from 0.383 to 0.455 relative to the unfiltered UCS training set.
For GCS, the same filter improves EM from 0.457 to 0.482 and $\Delta$-JS from 0.578 to 0.631, while maintaining an O-ES of 0.735. 
These gains indicate that irreducibility identifies samples that provide clearer supervision for exact execution behavior and context-patch consistency.

The low-irreducibility slices show the complementary effect.
Training exclusively on samples with $\widehat{R}_{64} < 0.5$ reduces performance in both synthesis regimes.
Under UCS, the low-irreducibility slice decreases EM to 0.269, O-ES to 0.476, and $\Delta$-JS to 0.141.
Under GCS, the corresponding scores are 0.359, 0.616, and 0.392.
These reductions suggest that low-irreducibility samples often contain redundant, masked, or behaviorally inert arguments that weaken the association between input structure and execution artifacts.

The strongest results occur when grammar constraints and irreducibility filtering are combined.
GCS with $\widehat{R}_{64} \geq 0.5$ achieves the best EM and $\Delta$-JS, while matching the best O-ES.
This pattern supports the conclusion that the two mechanisms address complementary effects.
Grammar constraints reduce the generation of invalid or degenerate arguments before execution.
Irreducibility filtering then selects samples whose emitted arguments are behaviorally necessary after execution.
Together, these mechanisms produce training data that is both syntactically admissible and empirically informative for execution-free shell simulation.

%% file: content/5_discussion.tex
\section{Discussion}
\label{sec:discussion}
The results presented in \Cref{sec:eval} demonstrate that grammar-constrained synthesis and irreducibility filtering improve execution-annotated shell input samples through complementary mechanisms.
Grammar-constrained synthesis improves the generated corpus by restricting arguments to command-specific grammar languages before execution.
This reduces invalid argument actions, repeated usage errors, and early failure modes that can obscure the contribution of later arguments.
As a result, GCS produces more inputs that reach utility-specific execution paths.

The Irreducibility metric helps select samples with shell inputs whose arguments are empirically coupled to execution behavior.
This distinction is important because executability does not guarantee behavioral informativeness.
High-irreducibility samples provide clearer supervision because changes in the input are more likely to correspond to changes in observable outputs or latent state transitions.
The performance observed by models trained on samples with higher irreducibility suggests that this form of argument-effect attribution improves the utility of the training corpus, particularly for exact behavior prediction and context-patch consistency.

The strongest results are observed in models trained on grammar-constrained samples with higher irreducibility.
This suggests that behavioral necessity, in the form of irreducibility and syntactic admissibility within grammatical constraints, functions as a supplementary control mechanism.
Grammar constraints reduce the number of invalid or degenerate samples during synthesis, while irreducibility filtering identifies samples whose observed behavior depends on the constructed arguments.

Low-irreducibility samples may still represent common user behavior, stable command outputs, and realistic error modes.
They may also be useful for learning shell input invariances.
A deployment-oriented corpus should therefore preserve some low-irreducibility samples to model invariant behavior, while using high-irreducibility samples to strengthen learning of argument-sensitive execution behavior.
Irreducibility should thus be interpreted as a complementary signal rather than a universal quality criterion.

\paragraph{Limitations and scope.}
The grammar suite was constructed to provide a controlled and behaviorally diverse subset of the Linux command space, serving as a reproducible basis for evaluating grammar-constrained synthesis.
The suite covers 85\% of commands observed in attacker logs.
Of those unsupported, about 13\% are commands such as \texttt{sed} and \texttt{awk}, whose execution behavior is primarily driven by externally defined functionality that lies beyond the scope of command-line arguments.
The remaining 2\% of commands comprise shell commands that involve multi-system interaction and whose execution behavior cannot be reliably captured.
The 86 grammars presented in this work are portable across system contexts, with differences in $L(G_{x,c})$ for different system contexts $x$ arising from system-specific productions (e.g., filenames, users).
Future work should investigate learned or semi-automated grammar-construction methods that expand command coverage while preserving the context-dependent terminals required for system-specific execution modeling.

The experiments are conducted on static microVM environments operating in fixed system contexts and should be interpreted as measurements of fidelity within evaluated system contexts.
Generalizing across system contexts would require models that condition on and maintain the active system context in addition to a given shell input.
The context patches collected by ShIOEnv provide a mechanism for maintaining such a representation.
Future work should design such architectures using datasets collected across diverse system contexts.
The focus of this work is on dataset construction, in which we use execution-simulation performance to measure dataset utility.

%% file: content/6_conclusion.tex
\section{Conclusion}
\label{sec:conclusion}
Execution-free shell modeling requires datasets that link shell inputs to both observable execution records and latent state transitions within a specified system context.
This paper introduced ShIOEnv, a Gymnasium-compatible data-generation environment for constructing system-specific, execution-annotated Bash shell inputs.
ShIOEnv executes a given input in a given system context and records stdout, stderr, exit status, and structured context patches.
ShIOEnv addresses the lack of system-specific execution annotations in existing shell corpora and provides a reproducible mechanism to construct supervised data for command-behavior modeling.

We introduced behavioral irreducibility as a data-quality signal for estimating whether input arguments are necessary for the observed execution behavior.
A fixed-budget Monte Carlo estimator provides a practical approximation to exhaustive ablation, enabling irreducibility to be used for corpus-scale analysis and filtering.
We also introduced grammar-constrained synthesis, which restricts argument generation to command-specific context-free grammars and reduces invalid, redundant, and error-dominated argument combinations.

We synthesized 2.1M execution-annotated samples spanning 86 Linux utilities.
The empirical evaluation shows that grammar-constrained synthesis produces shell inputs with higher behavioral irreducibility and improves execution-free shell simulation.
Models trained on grammar-constrained samples outperform those trained on unconstrained samples and on adapted NL2CMD samples.
Irreducibility filtering yields additional gains, particularly for exact behavior prediction and context-patch consistency.
These results indicate that syntactic admissibility and behavioral necessity address complementary sources of noise in execution-annotated shell data.

The results support the use of ShIOEnv as both a data generator and an evaluation instrument for system-specific command modeling.
The environment, grammars, execution artifacts, and generated datasets provide a framework for shell input-behavior mapping, data-quality filtering, and execution-free interaction.
Future work should extend grammar coverage to more expressive utilities, incorporate explicit environment-state conditioning, and study longer-horizon interaction settings in which context patches are used to update the modeled system state for multi-input sessions.
We release ShIOEnv, command grammars, and the generated datasets to support further research on data-centric methods for system-specific CLI modeling.


%% file: content/7_appendix.tex
\appendix

\section{Dataset Details}
\label{sec:append_dataset}
This appendix documents the execution-annotated datasets produced by ShIOEnv.
We separate datasets by synthesis source: adapted NL2CMD, unconstrained synthesis (UCS), and grammar-constrained synthesis (GCS).
The adapted NL2CMD dataset was bootstrapped from 44.9K samples and modified to include arguments specific to the system context in which they will be executed.
The released corpus contains 2.1M samples across all sources.
The evaluations performed in \Cref{sec:eval} use 500K samples per data source for controlled comparison.

\subsection{Dataset Format and Fields}
\label{subsec:dataset_format}

\begin{table}[ht]
    \centering
    \caption{ShIOEnv dataset fields}
    \label{tab:shioenv_dataset}
    \begin{tabular}{l l l}
    \toprule
    Field & Type & Description \\
    \midrule
    \texttt{session\_id} & Integer & Unique sample identifier \\
    \texttt{input} & String & Constructed input sequence \\
    \texttt{input\_args} & List & Input separated into argument units \\
    \texttt{exit\_code} & Integer & Exit code returned by the executed input \\
    \texttt{output} & String & Observed stdout and stderr from executing the input \\
    \texttt{context\_patch} & String & Latent changes to the abstract system environment \\
    \texttt{irreducibility} & Float & Length-weighted sub-input contrast score \\
    \bottomrule
    \end{tabular}
\end{table}

Each data sample is stored as a JSON object whose fields are enumerated in \Cref{tab:shioenv_dataset}.
The \texttt{session\_id} field contains an identifier for the sampled input.
The \texttt{input} field contains the final observed input emitted by the environment, including the initial working-directory command \texttt{cd <cwd>;}.
The \texttt{input\_args} field contains the input separated into argument units that can be removed without breaking syntactic validity.
Some arguments contain \texttt{<ns>} markers, denoting a no-space boundary.
If two arguments contain \texttt{<ns>} on the same boundary, they are concatenated.
For example, the input \texttt{"ls -laT 32"} is represented as \texttt{[ls, -l<ns>, <ns>a<ns>, <ns>T 32]}.
We use this representation to support argument-level synthesis and contrastive ablation while preserving shell-compatible input strings.

The \texttt{exit\_code} field contains the exit code returned by the last executed input in the sequence.
The \texttt{output} field contains the stdout and stderr produced by execution.
The \texttt{context\_patch} field contains latent changes to the executing environment, stored as a list of condensed RFC 6902-style object patches to an abstract environment context object, as shown in \Cref{fig:ctx_patch} \citep{rfc6902}.
The \texttt{irreducibility} field contains the length-weighted sub-input contrast score defined in \Cref{subsec:rew_signal}.
The \texttt{output} and \texttt{context\_patch} fields serve as prediction targets during evaluation.

\begin{figure}[ht]
    \centering
    \begin{minipage}{\textwidth}
    \begin{lstlisting}[language=Python,
    linewidth=0.6\textwidth,
    basicstyle=\ttfamily,
    frame=single,
    backgroundcolor=\color{lightgray!20},
    keywordstyle=\color{blue}\bfseries,
    commentstyle=\color{green!50!black}\itshape,
    stringstyle=\color{orange}]
["a", "/path", value]   # add
["=", "/path", value]   # replace
["r", "/path"]          # remove
["m", "/from", "/to"]   # move
    \end{lstlisting}
    \end{minipage}
    \caption{Context patch representation}
    \label{fig:ctx_patch}
\end{figure}

\begin{figure}[htb]
\centering    
\begin{subfigure}{0.9\textwidth}
    \centering
    \includegraphics[width=\textwidth]{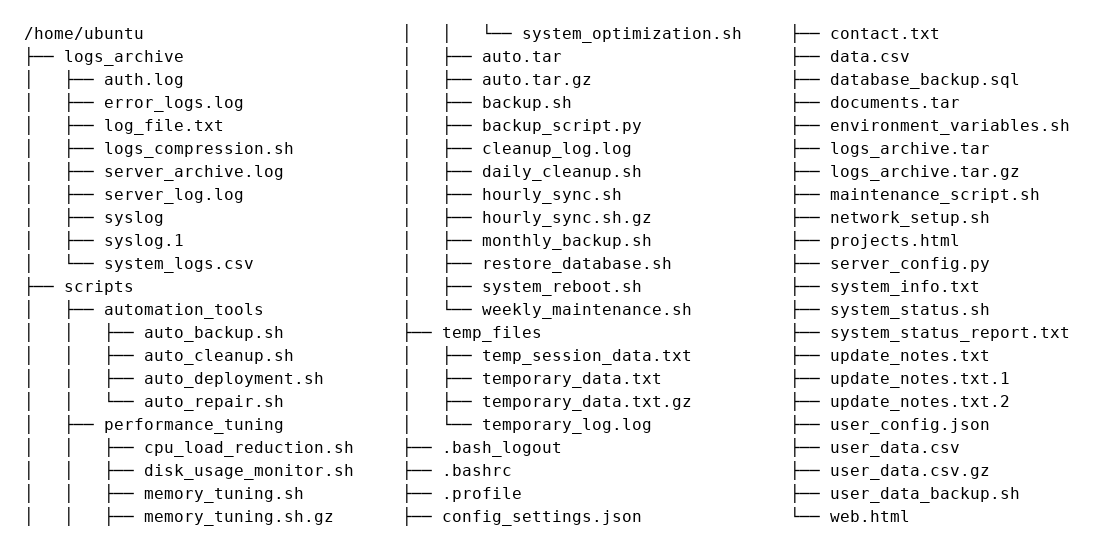}
    \caption{Ubuntu 24.04 filesystem}
    \label{subfig:ubuntu24_fs}
    \end{subfigure}
\hfill
\begin{subfigure}{0.95\textwidth}
    \centering
    \includegraphics[width=\textwidth]{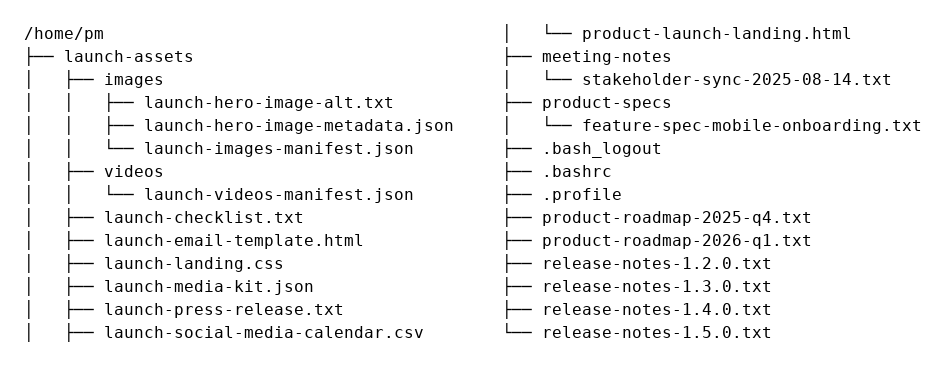}
    \caption{Ubuntu 22.04 filesystem}
    \label{subfig:ubuntu22_fs}
    \end{subfigure}
\hfill
\begin{subfigure}{0.95\textwidth}
    \centering
    \includegraphics[width=\textwidth]{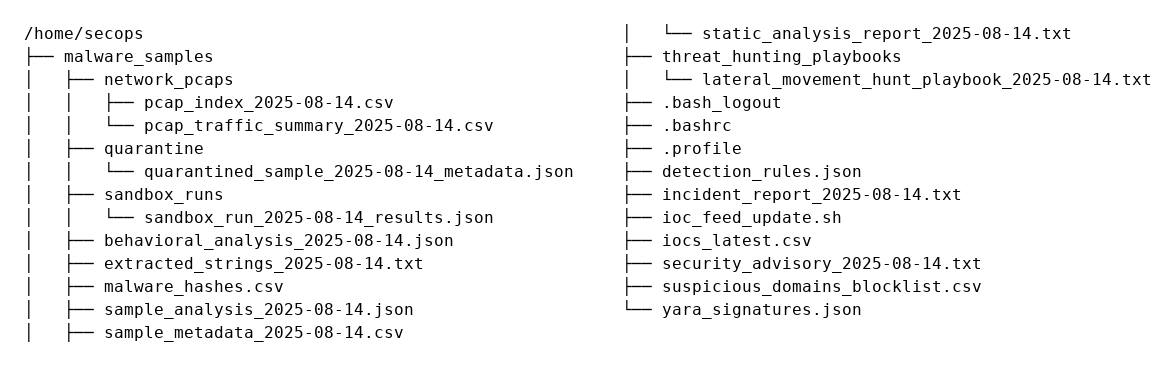}
    \caption{Debian 12 filesystem}
    \label{subfig:debian12_fs}
    \end{subfigure}

    \caption{System contexts used in evaluation. The Ubuntu 22.04 and Debian 12 file systems are only used for system context ablations reported in \Cref{tab:multi_sys}.}
    \label{fig:fs_tree}
\end{figure}

\section{Environment Details}
\label{sec:append_env}
Internally, ShIOEnv alternates between two phases: construction, during which the agent appends textual arguments to the current command, and execution, during which the constructed session is run from a fresh microVM snapshot to collect behavioral feedback.

\subsection{State Representation}
The environment is fully observable, wherein the state representation and the observations returned to the agent are equivalent.
Each state observation is an immutable, padded 2‑D tuple of UTF‑8 strings $(\Sigma^{\le 64})^{H_G\times H_L}$, where $\Sigma^{\le 64}$ denotes strings of at most 64 characters.
The first dimension contains each input in a session, comprised of $L$ or fewer arguments.
Empty strings serve as argument padding, ensuring the tensor maintains a fixed shape.
At reset, the first row is pre-initialized with $(\texttt{cd},c_{0})$ where $c_{0}$ is a starting directory drawn uniformly from a user‑defined set.
In the experiments conducted in \Cref{sec:eval}, $c_0$ is set to the home directory for each system context shown in \Cref{fig:fs_tree}.

Two internal counters $(h,l)$ track the agent's position inside the tensor.
Two horizon parameters cap the episode: the maximum number of commands per session ($H_\mathrm{G}$) and the maximum number of arguments per command ($H_\mathrm{L}$).
When either counter reaches its respective bound, the environment sets Gym's \texttt{truncate} flag to \texttt{True} and the \texttt{done} flag to \texttt{True} when $G$ is reached.

\subsection{Action Representation}
Each action is passed as a three-tuple:
\[
\langle \texttt{input\_addition:}s, \texttt{exec\_action:}e, \texttt{new\_global:}n \rangle,
\]
where $s\in\Sigma^{\le 64}$ describes an appending argument and $e,n\in{0,1}$ are boolean flags for execution and argument placement, respectively.
If $e=0$, the string $s$ is appended to the current command ($n=0$) or starts a new command ($n=1$).
If $e=1$, the current session is executed or concluded, and $s$ must be empty.

The environment imposes validity constraints on the actions provided.
The boolean flags are mutually exclusive.
Violations such as providing an \texttt{input\_addition} while $e=1$, or setting both Boolean flags to \texttt{True}, yield an immediate reward of $-10$ and terminate the episode.
Attempting to exceed either horizon sets done to \texttt{True}.

\begin{algorithm}[htb]
\caption{Adaptive noise threshold $\beta$}
\label{alg:beta}
\begin{algorithmic}[1]

\REQUIRE command sequence $S$; sample $N$; $O \gets \{\}$
\FOR{$i \gets 1$ to $N$}
    \STATE $O \gets O \cup \text{exec}(S)$
\ENDFOR
\STATE $\mathcal{P} \gets \bigl\{\operatorname{sim}(o_i, o_j)\,\bigl|\bigr.\, 1 \le i < j \le N\bigr\}$ \COMMENT{all $\genfrac{(}{)}{0pt}{}{N}{2}$ pair‑wise similarities}
\STATE $\mu \gets \operatorname{mean}(\mathcal{P})$
\STATE $\sigma \gets \operatorname{stdev}(\mathcal{P})$
\STATE $\beta \gets \min\!\bigl(0.95,\; \mu - 2\sigma\bigr)$
\RETURN $\beta$
\end{algorithmic}
\end{algorithm}

\subsection{Action and Execution Handling}
No executions take place in the environment until a sequence is complete and an executing action is received ($e=1$).
The environment then collects execution behaviors for the full input and each of its sub-inputs within the prescribed sampling budget.
Helper routines record the command's standard output, exit status, and a structured snapshot of the system context (working directory, directory contents, environment variables, user groups, shell options, resource limits, and firewall rules) before and after execution to compute a context change patch.
These procedures document the value of each system feature both prior to and following execution, creating a context patch for each that change.
We use the produced observations in the presented sub-input contrasts.
When a sub-input produces observations equivalent to those of the full input, this signifies the removal of redundant arguments.

We classify behaviors as identical if the exit codes and post-execution context patch match and if the normalized edit similarity of the observable output exceeds a noise threshold $\beta$.
$\beta$ is given as the mean pairwise edit similarities of $N$ identical executions, reduced by two standard deviations.
For our experiments, we collect $N=3$ traces to compute $\beta$, as given in \Cref{alg:beta}.
This threshold accounts for deviations arising from uncontrollable variables, such as host load or timestamps.
If the variance in observations is high, the threshold for edit-similarity-based redundancy will be low, requiring a greater content difference to safely classify an output as changed.
We find this over-punishment tolerable, as underestimating irreducibility ensures that fully irreducible inputs remain so.

\begin{figure}[htb]
    \centering
    \begin{minipage}{\linewidth}
\begin{verbatim}
You are emulating a Linux command-line interface for a user to 
interact with. Features of this emulated environment are as follows:
{
'arch': 'x86_64',
'kernel': '6.8',
'cpus': '1',
'mem': '216M',
'os_family': 'ubuntu-24',
'cwd': '/home/ubuntu',
...
'env': {'HOME': '/',
        ...
    },
'fs': {'.': {
          'group': 'ubuntu',
          'owner': 'ubuntu',
          'perms': 'drwxrwxr-x',
          'size': 4096,
          'time': '19:43',
          ...},
          ...
    },
 ...
 }
For each input, respond with:
1. An OUT segment containing exactly the text that would 
be printed to stdout 
or stderr, enclosed as <OUT>...</OUT>.
2. A DIFF segment containing a JSON-encoded compact RFC6902-style patch 
representing the changes to the abstract environment, enclosed as 
<DIFF>...</DIFF>.

The DIFF uses the following compact encoding (a JSON list of operations):
  ["a", "/path", value]   # add
  ["=", "/path", value]   # replace
  ["r", "/path"]          # remove
  ["m", "/from", "/to"]   # move

Always respond in the exact format:
<OUT> OUTPUT_TEXT </OUT> <DIFF> JSON_PATCH </DIFF>
\end{verbatim}
  \end{minipage}
    \caption{Instruction template provided for instruction-tuned system modeling}
    \label{fig:append_llm}
\end{figure}

\section{GPT Model Input}
\label{sec:append_llm}
We evaluate models trained on datasets curated using the proposed data synthesis methods against existing methods for execution-free CLI interaction.
Among these methods is the use of a prompted large language model.
For this comparison, we pass each evaluated input along with instructions to provide execution responses and output for modeling execution behavior.
A truncated version of these instructions is given in \cref{fig:append_llm}.